\documentclass[english]{article}
\usepackage[T1]{fontenc}
\usepackage[utf8]{inputenc}
\usepackage{wrapfig}
\usepackage{mathtools}
\usepackage{url}
\usepackage{amsmath}
\usepackage{amsthm}
\usepackage{amssymb}
\usepackage{graphicx}

\makeatletter
\usepackage[final,nonatbib]{neurips_2024}

\usepackage[utf8]{inputenc} 
\usepackage[T1]{fontenc}    
\usepackage{hyperref}       
\usepackage{url}            
\usepackage{booktabs}       
\usepackage{amsfonts}       
\usepackage{nicefrac}       
\usepackage{microtype}      
\usepackage{xcolor}         

\author{%
  Bastian Epping$^1$, Alexandre René$^1$, Moritz Helias$^{2,3}$, Michael T.~Schaub$^1$ \\
  $^1$RWTH Aachen University, Aachen, Germany \\
  $^2$Department of Physics, RWTH Aachen University, Aachen, Germany \\
  $^3$Institute for Advanced Simulation (IAS-6), Computational and Systems Neuroscience, \\ Jülich Research Centre, Jülich, Germany \\
  \texttt{epping@cs.rwth-aachen.de, rene@cs.rwth-aachen.de,} \\
  \texttt{m.helias@fz-juelich.de, schaub@cs.rwth-aachen.de} \\
}

\makeatother

\usepackage{babel}
\begin{document}
\title{Graph Neural Networks Do Not Always Oversmooth}
\maketitle
\begin{abstract}
Graph neural networks (GNNs) have emerged as powerful tools for processing
relational data in applications. However, GNNs suffer from the problem
of oversmoothing, the property that features of all nodes exponentially
converge to the same vector over layers, prohibiting the design of
deep GNNs. In this work we study oversmoothing in graph convolutional
networks (GCNs) by using their Gaussian process (GP) equivalence in
the limit of infinitely many hidden features. By generalizing methods
from conventional deep neural networks (DNNs), we can describe the
distribution of features at the output layer of deep GCNs in terms
of a GP: as expected, we find that typical parameter choices from
the literature lead to oversmoothing. The theory, however, allows
us to identify a new, non-oversmoothing phase: if the initial weights
of the network have sufficiently large variance, GCNs \emph{do not}
oversmooth, and node features remain informative even at large depth.
We demonstrate the validity of this prediction in finite-size GCNs
by training a linear classifier on their output. Moreover, using the
linearization of the GCN GP, we generalize the concept of propagation
depth of information from DNNs to GCNs. This propagation depth diverges
at the transition between the oversmoothing and non-oversmoothing
phase. We test the predictions of our approach and find good agreement
with finite-size GCNs. Initializing GCNs near the transition to the
non-oversmoothing phase, we obtain networks which are both deep and
expressive.
\end{abstract}

\section{Introduction}

Graph neural networks (GNNs) reach state of the art performance in
diverse application domains with relational data that can be represented
on a graph, transferring the success of machine learning to data on
graphs \cite{wu_comprehensive_2021,hamilton_graph_2020,ma_deep_2021,defferrard_convolutional_2016}.
Despite their good performance, GNNs come with the limitation of \emph{oversmoothing},
a phenomenon where node features converge to the same state exponentially
fast for increasing depth \cite{rusch_survey_2023,wu_demystifying_2023,oono_graph_2021,cai_note_2020}.
Consequently, only shallow networks are used in practice \cite{kipf_semi-supervised_2017,alon_bottleneck_2021}.
In contrast, it is known that the depth (i.e. the number of layers)
is key to the success of deep neural networks (DNNs) \cite{poole_exponential_2016,raghu_expressive_2017}.
While for conventional DNNs shallow networks are proven to be highly
expressive \cite{cybenko_approximation_1989}, in practice deep networks
are much easier to train and are thus the commonly used architectures
\cite{saxe_exact_2014}. Furthermore, in most GNN architectures each
layer only exchanges information between neighboring nodes. Deep GNNs
are therefore necessary to exchange information between nodes that
are far apart in the graph \cite{dwivedi_long_2022}. In this study,
we investigate oversmoothing in graph convolutional networks (GCNs)
\cite{kipf_semi-supervised_2017}.

To study the effect of depth, we consider the propagation of features
through the network: given some input $\boldsymbol{x}_{\alpha}^{(0)}$,
each intermediate layer $l$ produces features $\boldsymbol{x}_{\alpha}^{(l)}$
which are fed to the next layer. We follow the same approach that
has successfully been employed in previous work to design trainable
DNNs \cite{schoenholz_deep_2017}: consider two nearly identical inputs
$\boldsymbol{x}_{\alpha}^{(0)}$ and $\boldsymbol{x}_{\beta}^{(0)}$
and ask whether the intermediate features $\boldsymbol{x}_{\alpha}^{(l)}$
and $\boldsymbol{x}_{\beta}^{(l)}$ become more or less similar as
a function of depth $l$. In the former case, the inputs may eventually
become indistinguishable. In the latter case, the inputs become less
similar over layers: the distance between them increases over layers
\cite{poole_exponential_2016,schoenholz_deep_2017} until eventually
it is bounded by the non-linearities of the network. The distance
then typically converges to a fixed value determined by the network
architecture, independent of the inputs.

One can therefore identify two phases: One says that a network is
\emph{regular} if two inputs eventually converge to the same value
as function of $l$; conversely, one says that a network is \emph{chaotic}
if two inputs remain distinct for all depths \cite{molgedey_suppressing_1992}.
Neither phase is ideal for training deep networks since in both cases
all the information from the inputs is eventually lost; the typical
depth at which this happens is called the \emph{information propagation
depth}. However, this propagation depth diverges at the transition
between the two phases, allowing information -- in principle --
to propagate infinitely deep into the network. While these results
are calculated in the limit of infinitely many features in each hidden
layer, the information propagation depth has been found to be a good
indicator of how deep a network can be trained \cite{schoenholz_deep_2017}.
A usual approach for conventional DNNs is thus to initialize them
at the transition to chaos. Indeed, Schoenholz et al.~\cite{schoenholz_deep_2017}
were able to use this approach to train fully-connected, feedforward
networks with hundreds of layers. Similar methods have recently been
adapted to the study of transformers, successfully predicting the
best hyperparameters for training \cite{cowsik_geometric_2024}.

In this work we address the oversmoothing problem of GCNs by extending
the framework described above in the limit of infinite feature dimensions
from DNNs to GCNs: here the two different inputs $\boldsymbol{x}_{\alpha}^{(0)}$
and $\boldsymbol{x}_{\beta}^{(0)}$ correspond to the input features
on two nodes, labeled $\alpha$ and $\beta$. The mixing of information
across different nodes implies that output features on node $\alpha$
depend on input features on node $\beta$ and vice versa. Thus it
is not possible to look at the distance of $\boldsymbol{x}_{\alpha}^{(l)}$
and $\boldsymbol{x}_{\beta}^{(l)}$ independently for each pair $\alpha$
and $\beta$ as in the DNN case. Rather, one has to solve for the
distance between each distinct pair of nodes in the graph simultaneously:
the one dimensional problem for DNNs thus becomes a multidimensional
problem for GCNs. However, by linearizing the multidimensional GCN
dynamics, we can generalize the notion of information propagation
depth to GCNs: instead of being a single value, we find that a given
GCN architecture comes with a set of potentially different information
propagation depths, each corresponding to one eigendirection of the
linearized dynamics of the system.

This approach allows us to extend the concept of a regular and a chaotic
phase to GCNs: in the regular phase, which describes most of the GCNs
studied in the current literature, distances between node features
shrink over layers and exponentially attain the same value. We therefore
call this the oversmoothing phase. On the other hand, if one increases
the variance of the weights at initialization, it is possible to transition
into the chaotic phase. In this phase, distances between node features
converge to a fixed but finite distance at infinite depth. The convergence
point is fully determined by the underlying graph structure and the
hyperparameters of the GCN and may differ for different pairs of nodes.
GCNs initialized in this phase thus do not suffer from oversmoothing.
We find that the convergence point is informative about the topology
of the underlying graph and may be used for node classification with
GCNs of more than $1,000$ layers. Near the transition point, GCNs
at large depth offer a trade-off between feature information and information
contained in the neighborhood relation of the graph. We test the predictions
of this theory and find good agreement in comparison to finite-size
GCNs applied to the contextual stochastic block model \cite{deshpande_contextual_2018}.
On the citation network Cora \cite{kipf_semi-supervised_2017} we
reach the performance reported in the original work by Kipf and Welling
\cite{kipf_semi-supervised_2017} beyond $100$ layers. Our approach
applies to graphs with arbitrary topologies, depths, and non-linearities.

\section{Related Work}

Oversmoothing is a well-known challenge within the GNN literature
\cite{li_deeper_2018,rusch_survey_2023}. On the theoretical side,
rigorous techniques have been used to prove that oversmoothing is
inevitable. The authors in \cite{oono_graph_2021} show that GCNs
with the ReLU non-linearity exponentially lose expressive power and
only carry information of the node degree in the infinite layer limit.
While the authors notice that their upper bound for oversmoothing
does not hold for large singular values of the weight matrices, they
do not identify a non-oversmoothing phase in their model. These results
have been extended in \cite{cai_note_2020} to handle non-linearities
different from ReLU. Also graph attention networks have been proven
to oversmooth inevitably \cite{wu_demystifying_2023}. Their proof,
however, makes assumptions on the weight matrices which, as we will
show, exclude networks in the chaotic, non-oversmoothing phase.

On the applied side, a variety of heuristics have been developed to
mitigate oversmoothing \cite{zhao_pairnorm_2020,chen_simple_2020,chen_measuring_2020,luan_training_2022,song_ordered_2023,huang_you_2024}.
E.g., the authors in \cite{zhao_pairnorm_2020} introduce a normalization
layer which can be added to a variety of deep GNNs to make them trainable.
Another approach is to introduce residual connections and identity
mappings, directly feeding the input to layers deep in the network
\cite{chen_simple_2020,wu_non-asymptotic_2023}. Other studies suggest
to train GNNs to a limited number of layers to obtain the optimal
amount of smoothing \cite{keriven_not_2022,wu_non-asymptotic_2023}.
The recent review \cite{jin_atnpa_2024} proposes a unified view to
order existing heuristics and guide further research. While these
heuristics improve performance at large depths, they also add to the
complexity of the model and impose design choices. Our approach, on
the other hand, explains why increasing the weight variance at initialization
is sufficient to prevent oversmoothing.

\section{Background}

\subsection{\label{subsec:Network-architecture}Network architecture}

In this paper we study a standard graph convolutional network (GCN)
architecture \cite{kipf_semi-supervised_2017} with an input feature
matrix $\boldsymbol{X}^{(0)}\in\mathbb{R}^{N\times d_{0}}$, where
$N$ is the number of nodes in the graph and $d_{0}$ the number of
input features. Bold symbols throughout represent vector or matrix
quantities in feature space. The structure of the graph is represented
by a shift operator $A\in\mathbb{R}^{N\times N}$. We write the features
of the network’s $l$-th layer as $\boldsymbol{X}^{(l)}\in\mathbb{R}^{N\times d_{l}}$;
they are computed recursively as
\begin{align}
\boldsymbol{X}^{(l)} & =\phi(A\boldsymbol{X}^{(l-1)}\boldsymbol{W}^{(l)\top}+1\boldsymbol{b}^{(l)\top})\,,\label{eq:basic_eq_gnn}
\end{align}
with $\phi$ an elementwise non-linear activation function, $\boldsymbol{b}^{(l)}$
an optional bias term, weight matrices $\boldsymbol{W}^{(l)}\in\mathbb{R}^{d_{l}\times d_{l-1}}$,
and $1\in\mathbb{R}^{N}$ a vector of all ones. We note that many
GNN architectures studied in the literature are unbiased, which can
be recovered by setting $\boldsymbol{b}^{(l)}$ to zero. We use a
noisy linear readout, so that the output of the network is given by

\begin{equation}
\boldsymbol{Y}=A\boldsymbol{X}^{(L)}\boldsymbol{W}^{(L+1)\top}+1\boldsymbol{b}^{(L+1)\top}+\boldsymbol{\epsilon}\,,
\end{equation}
with $\boldsymbol{\epsilon}\in\mathbb{R}^{N\times d_{L+1}}$ being
independent Gaussian random variables: $\epsilon_{\alpha,i}\overset{\mathrm{i.i.d.}}{\sim}\mathcal{N}(0,\sigma_{ro}^{2})$.
The readout noise $\boldsymbol{\epsilon}$ is included both to promote
robust outputs and to prevent numerical issues in the matrix inversion
in Equations~(\ref{eq:general-GP-posterior-mean}) and (\ref{eq:general-GP-posterior-variance}).
We use $d_{L+1}$ to denote the dimension of outputs.

For the following it will be useful to consider the activity of individual
nodes. To avoid ambiguity in the indexing, we use lower Greek indices
for nodes and upper Latin indices for layers. We thus rewrite (\ref{eq:basic_eq_gnn})
as
\begin{align}
\boldsymbol{x}_{\alpha}^{(l)} & =\phi(\boldsymbol{h}_{\alpha}^{(l)})\,,\label{eq:gcn_dynamics_feature_from_preactivation}\\
\boldsymbol{h}_{\alpha}^{(l)} & =\sum_{\beta}A_{\alpha\beta}\boldsymbol{W}^{(l)}\boldsymbol{x}_{\beta}^{(l-1)}+\boldsymbol{b}^{(l)}\,,\label{eq:gcn_dynamics_preact_from_features}\\
\boldsymbol{y}_{\alpha} & =\boldsymbol{h}_{\alpha}^{(L+1)}+\boldsymbol{\epsilon}_{\alpha}\,,
\end{align}
where $\boldsymbol{x}_{\alpha}^{(l)}\in\mathbb{R}^{d_{l}}$ is the
feature vector of node $\alpha$ in layer $l$ and $\boldsymbol{y}_{\alpha}\in\mathbb{R}^{d_{L+1}}$
the network output for node $\alpha$. The values $\boldsymbol{h}_{\alpha}^{(l)}\in\mathbb{R}^{d_{l}}$
are linear functions of the features $\boldsymbol{x}_{\beta}^{(l-1)}$
and represent the input to the activation functions; we therefore
refer to them as preactivations. The non-linearity $\phi(x)$ is applied
elementwise to the preactivations $\boldsymbol{h}_{\alpha}^{(l)}$.
While we leave $\phi(x)$ general for the development of the theory,
we use $\phi(x)=\mathrm{erf}(\frac{\sqrt{\pi}}{2}x)$ for the experiments
in Section~\ref{sec:Results}; this choice allows us to carry out
certain integrals analytically. The scaling factor in the $\mathrm{erf}$
is chosen such that $\frac{\partial\phi}{\partial x}(0)=1$. We use
independent and identical Gaussian priors for all weight matrices
and biases, $W_{ij}^{(l)}\overset{\mathrm{i.i.d.}}{\sim}\mathcal{N}(0,\frac{\sigma_{w}^{2}}{d_{l}})$
and $b_{i}^{(l)}\overset{\mathrm{i.i.d.}}{\sim}\mathcal{N}(0,\sigma_{b}^{2})$
with $W_{ij}^{(l)}$ and $b_{i}^{(l)}$ being the matrix or vector
entries of $\boldsymbol{W}^{(l)}$ and $\boldsymbol{b}^{(l)}$, respectively.
As a shift operator, we choose
\begin{equation}
A=\mathbb{I}-\frac{g}{d_{\mathrm{max}}}(D-\mathcal{A})\,,\label{eq:shift-operator-definition}
\end{equation}
where $\mathcal{A}$ is the adjacency matrix, $\mathbb{I}$ the identity
in $\mathbb{R}^{N\times N}$, $D_{\alpha\beta}=\delta_{\alpha\beta}\sum_{\gamma}\mathcal{A_{\alpha\gamma}}$
is the degree matrix and $d_{\mathrm{max}}$ is the maximal degree.
The parameter $g\in(0,1)$ allows us to weigh the off-diagonal elements
compared to the diagonal ones. By construction the shift operator
is row-stochastic, which means that it has constant sums over columns
$\sum_{\beta}A_{\alpha\beta}=1$. We will make use of this property
in our analysis in Section~\ref{subsec:The-non-oversmoothing-phase}.
The generalization to non-stochastic shift operators will be shortly
addressed later.

\subsection{Gaussian process equivalence of GCNs}

In a classic machine learning setting, such as classification, one
draws random initial values for all parameters and subsequently trains
the parameters by optimizing the weights and biases to minimize a
loss function. This learned parameter set is then used to classify
unlabeled inputs. In this paper we take a Bayesian point of view in
which the network parameters are random variables, inducing a probability
distribution over outputs which becomes Gaussian in the limit of infinitely
many features. Thus infinitely wide neural networks are equivalent
to Gaussian processes (GPs) \cite{neal_priors_1994,williams_computing_1996,rasmussen_gaussian_2006}.
In the study of DNNs this is a standard approach, yielding results
which empirically hold also for finite-size networks trained with
gradient descent \cite{schoenholz_deep_2017}.

In previous work \cite{ng_bayesian_2018,hu_infinitely_2020,niu_graph_2023}
it has been shown that also the GCN architecture described in Section~\ref{subsec:Network-architecture}
is equivalent to a GP in the limit of infinite feature space dimensions,
$d_{l}\to\infty$ for all hidden layers $l=1,\dots,L$, while input
and readout layer still have tunable, finitely many features. In the
GP description, all features are Gaussian random variables with zero
mean and identical prior variance in each feature dimension. The description
of the GCN thus reduces to a multivariate normal,
\begin{align}
H^{(l)} & \sim\mathcal{N}(0,K^{(l)})\,,
\end{align}
where $H^{(l)}$ is the vector of hidden node features of layer $l$,
$H^{(l)}=(h_{0}^{(l)},h_{1}^{(l)},\dots,h_{N}^{(l)})^{\top}$ under
the prior distribution of weights and biases. The covariance matrices
$K^{(l)}\in\mathbb{R}^{N\times N}$ are determined recursively: knowing
that the $h_{\alpha}^{(l)}$ follow a zero-mean Gaussian with covariance
$\langle h_{\delta}^{(l)}h_{\gamma}^{(l)}\rangle=K_{\delta\gamma}^{(l)}$,
we define
\begin{align}
C_{\gamma\delta}^{(l)} & =\Big\langle\phi\Big(h_{\gamma}^{(l)}\Big)\phi\Big(h_{\delta}^{(l)}\Big)\Big\rangle_{h_{\gamma}^{(l)},h_{\delta}^{(l)}}\,.\label{eq:GCNGP_expectation_value}
\end{align}
For simplicity we use $\phi(x)=\mathrm{erf}(\frac{\sqrt{\pi}}{2}x)$
for which Equation~(\ref{eq:GCNGP_expectation_value}) can be evaluated
analytically; see Appendix~\ref{app:Analytical-solution-for} for
details. It follows from (\ref{eq:gcn_dynamics_feature_from_preactivation})
that
\begin{align}
K_{\alpha\beta}^{(l+1)} & =\sigma_{b}^{2}+\sigma_{w}^{2}\,\sum_{\gamma,\delta}A_{\alpha\gamma}A_{\beta\delta}C_{\gamma\delta}^{(l)}\,,\label{eq:GCNGP_K_update}
\end{align}
as shown in \cite{ng_bayesian_2018,niu_graph_2023}.

In a semi-supervised node classification setting, we split the underlying
graph into $N^{\mathrm{test}}$ unlabeled test nodes and $N^{\mathrm{train}}$
labeled training nodes ($N^{\mathrm{test}}+N^{\mathrm{train}}=N$);
we correspondingly split the output random variable $Y_{i}\in\mathbb{R}^{N}$
for output dimension $i$ into $Y_{i}^{\star}\in\mathbb{R}^{N^{\mathrm{test}}}$
and $Y_{i}^{D}\in\mathbb{R}^{N^{\mathrm{train}}}$. Features on the
test nodes are predicted by conditioning on the values of the training
nodes: $p\bigl(Y_{i}^{\ast}=y_{i}^{\ast}\mid Y_{i}^{D}=y_{i}^{D}\bigr)\,.$
This leads to the following posterior for the unobserved labels (see
\cite{rasmussen_gaussian_2006,lee_deep_2018} for details):
\begin{align}
Y_{i}^{\star} & \sim\mathcal{N}(m_{i}^{\mathrm{GP}},K^{\mathrm{GP}})\,,\\
m_{i}^{\mathrm{GP}} & =K_{\star D}^{(L+1)}(K_{DD}^{(L+1)}+\mathbb{I}\sigma_{ro}^{2})^{-1}Y_{i}^{D}\,,\label{eq:general-GP-posterior-mean}\\
K^{\mathrm{GP}} & =K_{\star\star}^{(L+1)}-K_{\star D}^{(L+1)}(K_{DD}^{(L+1)}+\mathbb{I}\sigma_{ro}^{2})^{-1}(K_{\star D}^{(L+1)})^{\top}\,.\label{eq:general-GP-posterior-variance}
\end{align}
Here the $\star$ and $D$ indices represent test and training data,
respectively, i.e.~$K_{DD}\in\mathbb{R}^{N^{\mathrm{train}}\times N^{\mathrm{train}}}$
is the covariance matrix of outputs of all training nodes and $K_{\star D}\in\mathbb{R}^{N^{\mathrm{test}}\times N^{\mathrm{train}}}$
is the covariance between test data and training data. Finally, $\mathbb{I}$
is here the identity in $\mathbb{R}^{N^{\mathrm{train}}\times N^{\mathrm{train}}}$.

\subsection{Feature distance}

To measure and quantify how much a given GCN instance oversmoothes
we use the squared Euclidean distance between pairs of nodes, and
normalize by the number of node features $d_{l}$ so that the measure
stays finite in the GP limit $d_{l}\to\infty$. This allows us to
quantitatively test the predictions of our approach on the node-resolved
distances of features. To summarize the amount of oversmoothing across
the GCN, we also define the measure $\mu(X)$ as the average squared
Euclidean distance across all pairs of nodes:
\begin{align}
d(\boldsymbol{x}_{\alpha},\boldsymbol{x}_{\beta}) & =\frac{1}{d_{l}}||\boldsymbol{x}_{\alpha}-\boldsymbol{x}_{\beta}||_{2}^{2}=C_{\alpha\alpha}^{\prime}+C_{\beta\beta}^{\prime}-2C_{\alpha\beta}^{\prime}\,,\label{eq:2-distance-definition}\\
\mu(\boldsymbol{X}) & =\frac{1}{2N(N-1)}\sum_{\alpha=1}^{N}\sum_{\beta=\alpha+1}^{N}d(\boldsymbol{x}_{\alpha},\boldsymbol{x}_{\beta})\,.\label{eq:node-sim-definition}
\end{align}
Here $C_{\alpha\beta}^{\prime}=\frac{\boldsymbol{x}_{\alpha}\cdot\boldsymbol{x}_{\beta}}{d_{l}}$
is the normalized scalar product. We use the notation $C_{\alpha\beta}^{\prime}$
to avoid confusion with the expectation value $C_{\alpha\beta}$ defined
in the GCN GP (\ref{eq:GCNGP_expectation_value}). In the infinite
feature dimensions limit, the quantities $C_{\alpha\beta}^{\prime}$
in Equation~(\ref{eq:node-sim-definition}) converge to the GCN GP
quantities $C_{\alpha\beta}$ defined by (\ref{eq:GCNGP_expectation_value}).
In the following sections we will therefore use the $C_{\alpha\beta}$
as predictions for the $C_{\alpha\beta}^{\prime}$ of finite-size
GCNs. The normalization for $d(\boldsymbol{x}_{\alpha},\boldsymbol{x}_{\beta})$
and $\mu(\boldsymbol{X})$ can be interpreted as an average (squared)
feature distance, independent of the size of the graph and the number
of feature dimensions.

\section{\label{sec:Results}Results}

\subsection{Propagation depths}

We are interested in analyzing GCNs at large depth. We a priori assume
that at infinite depth the GCN converges to an equilibrium in which
covariances are static over layers $K_{\alpha\beta}^{(l)}\xrightarrow{l\to\infty}K_{\alpha\beta}^{\mathrm{eq}}$,
irrespective of whether the GCN is in the oversmoothing or the chaotic
phase. A posteriori we show that this assumption indeed holds. Since
the fixed point $K^{\mathrm{eq}}$ is independent of the input, a
GCN at equilibrium cannot use information from the input to make predictions
(although, as we will see, in the non-oversmoothing phase it can still
use the graph structure). In the following we analyze the equilibrium
covariance $K^{\mathrm{eq}}$ to which GCNs with different $\sigma_{w}^{2}$,
$\sigma_{b}^{2}$ and $A$ converge to, how they behave near this
equilibrium, and at which rate it is approached.

Close to equilibrium, the covariance matrix $K^{(l)}$ can be written
as a perturbation around $K_{\alpha\beta}^{\mathrm{eq}}$:
\begin{align}
K_{\alpha\beta}^{(l)} & =K_{\alpha\beta}^{\mathrm{eq}}+\Delta_{\alpha\beta}^{(l)}\,.\label{eq:linearized-covariance}
\end{align}
Under the assumption that the perturbation $\Delta_{\alpha\beta}^{(l)}$
is small, we can linearize the GCN GP
\begin{align}
\Delta_{\alpha\beta}^{(l+1)} & =\sum_{\gamma,\delta}H_{\alpha\beta,\gamma\delta}\Delta_{\gamma\delta}^{(l)}+\mathcal{O}((\Delta^{(l)})^{2})\,,\label{eq:linearized-gcngp}\\
H_{\alpha\beta,\gamma\delta} & =\sigma_{w}^{2}\sum_{\theta,\phi}\frac{1}{2}(1+\delta_{\gamma,\delta})A_{\alpha\theta}A_{\beta\phi}\frac{\partial C_{\theta\phi}}{\partial K_{\gamma\delta}}[K^{\mathrm{eq}}]\,,
\end{align}
where we use square brackets to denote the point around which we linearize.
The factor $\frac{1}{2}(1+\delta_{\gamma,\delta})$ is introduced
to correctly count on- and off-diagonal elements of the covariance
matrix, while the shift operators $A$ and the derivative $\frac{\partial C_{\theta\phi}}{\partial K_{\gamma\delta}}[K^{\mathrm{eq}}]$
originate from the message passing and the non-linearity $\phi$,
respectively. The latter would result in a Kronecker delta $\frac{\partial C_{\theta\phi}}{\partial K_{\gamma\delta}}[K^{\mathrm{eq}}]=\delta_{\theta\phi,\gamma\delta}$
for linear networks. The calculation for $H_{\alpha\beta,\gamma\delta}$
is done in detail in Appendix~\ref{app:Linearized-Gaussian-process}.

A conceptually similar linearization has been done in \cite{schoenholz_deep_2017}
for DNNs. In the DNN case, different inputs to the networks---which
correspond to input features on different nodes here---can be treated
separately, leading to decoupling of Equation~(\ref{eq:linearized-gcngp}).
The shift operator in the GCN dynamics, in contrast, couples features
on neighboring nodes -- the matrix $H_{\alpha\beta,\gamma\delta}$
is in general not diagonal.

We can still achieve a decoupling by interpreting Equation~(\ref{eq:linearized-gcngp})
as a matrix multiplication, if $\alpha\beta$ and $\gamma\delta$
are understood as double indices, and by finding the eigendirections
of the matrix $H\in\mathbb{R}^{N^{2}\times N^{2}}$. Taking the right
eigenvectors $V_{\alpha\beta}^{(i)}$ as basis vectors, we can decompose
the covariance matrix $\Delta_{\alpha\beta}^{(l)}=\sum_{i}\Delta_{i}^{(l)}V_{\alpha\beta}^{(i)}$
and thus obtain the overlaps $\Delta_{i}^{(l)}$ which evolve independently
over layers. If the fixed point $K^{\mathrm{eq}}$ is attractive,
all eigenvalues have absolute values smaller than one: $|\lambda_{i}|<1$.
This allows us to define the propagation depth $\xi_{i}\coloneqq-\frac{1}{\ln(\lambda_{i})}$
for each eigendirection, very similar to the DNN case \cite{schoenholz_deep_2017}.
In this form, the linear update equation (\ref{eq:linearized-gcngp})
simplifies to
\begin{align}
\Delta_{i}^{(l+d)} & =\lambda_{i}^{d}\Delta_{i}^{(l)}=\exp(-d/\xi_{i})\Delta_{i}^{(l)}\,,
\end{align}
thus decoupling the system. For details on the linearization and some
properties of the transition matrix $H$ refer to Appendix~\ref{app:Linearized-Gaussian-process}.

\subsection{\label{subsec:The-non-oversmoothing-phase}The non-oversmoothing
phase of GCNs}

In this section we establish the chaotic, non-oversmoothing phase
of GCNs, and show that this phase can be reached by simple tuning
of the weight variance $\sigma_{w}^{2}$ at initialization. We start
by noticing that a GCN is at a state of zero feature distance $\mu(\boldsymbol{X}^{(l)})=0$,
if the covariance matrix has constant entries, $K_{\alpha\beta}^{(l)}=k^{(l)}$:
Constant entries in $K_{\alpha\beta}^{(l)}$ imply that all preactivations
are the same, $\boldsymbol{h}_{\alpha}^{(l)}=\boldsymbol{h}_{\beta}^{(l)}$,
which in turn implies $C_{\alpha\beta}^{(l)}=c^{(l)}$ (by Equation~(\ref{eq:GCNGP_expectation_value}));
the latter is equivalent to features being the same, $\boldsymbol{x}_{\alpha}^{(l)}=\boldsymbol{x}_{\beta}^{(l)}$.
Due to our choice of the shift operator, the state of zero distance
(and thus of $K_{\alpha\beta}^{(l)}=k^{(l)}$) is always a fixed point.
Assuming that $C_{\alpha\beta}^{(l)}=c^{(l)}$, we obtain
\begin{align}
K_{\alpha\beta}^{(l+1)} & =\sigma_{b}^{2}+\sigma_{w}^{2}\underbrace{\sum_{\gamma}A_{\alpha\gamma}}_{=1}\underbrace{\sum_{\delta}A_{\beta\delta}}_{=1}c^{(l)}=k^{(l+1)}\,.\label{eq:GCNGP-oversmoothed-fixed-point}
\end{align}
In an overmoothing GCN, this fixed point is also attractive, meaning
that also pairs of feature inputs $\boldsymbol{x}_{\alpha}^{(0)}$,
$\boldsymbol{x}_{\beta}^{(0)}$ which initially have non-zero distance
$d(\boldsymbol{x}_{\alpha}^{(0)},\boldsymbol{x}_{\beta}^{(0)})\neq0$
(and thus $\mu(\boldsymbol{X}^{(0)})\neq0$) eventually converge to
the point of vanishing distance. The chaotic, non-oversmoothing phase
of a GCN is determined by the condition that this point of constant
covariance $K_{\alpha\beta}^{(l)}=k^{(l)}$ becomes unstable. More
formally, this can be written in terms of eigenvalues of the linearized
dynamics as
\begin{align}
\max\{|\lambda_{i}^{\mathrm{p}}|\} & \overset{?}{>}1\,.\label{eq:chaos-condition}
\end{align}
Here and in the following we will use the superscript $\mathrm{p}$
to denote that the linearization is done around the state of constant
covariance across nodes in both the oversmoothing and non-oversmoothing
phase. The propagation depth $\xi_{i}\coloneqq-\frac{1}{\ln(\lambda_{i})}$
diverges at the phase transition where one $\lambda_{i}$ approaches
$1$. Intuitively speaking, Equation~(\ref{eq:chaos-condition})
asks whether a small perturbation from the zero distance case diminishes
($\max\{|\lambda_{i}^{\mathrm{p}}|\}<1$), in which case the network
dynamics is regular, or grows ($\max\{|\lambda_{i}^{\mathrm{p}}|\}>1$),
in which case the network is chaotic and thus does not oversmooth.
The value of $\max\{|\lambda_{i}^{\mathrm{p}}|\}$ depends on the
choices of $A$, $\sigma_{w}^{2}$ and $\sigma_{b}^{2}$ (by the dependence
of $K^{\mathrm{eq}}$ on $\sigma_{b}^{2}$). In the following we will
concentrate on tuning $\sigma_{w}^{2}$ to reach the non-oversmoothing
phase.

\subsubsection{Complete graph}

\begin{figure}
\begin{centering}
\includegraphics[width=1\linewidth]{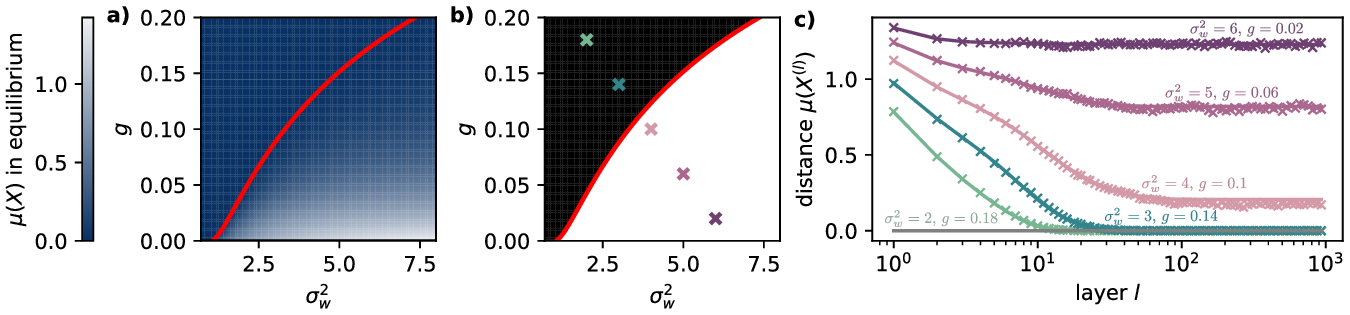}
\par\end{centering}
\centering{}\caption{\label{fig:toy-model-phase-diagram}Simulations and GP prior of a
GCN on a complete graph with $N=5$ nodes, shift operator $A_{\alpha\beta}=\frac{g}{N-1}+\delta_{\alpha\beta}(1-\frac{Ng}{N-1})$,
vanishing bias $\sigma_{b}^{2}=0$ and $\phi(x)=\mathrm{erf}(\frac{\sqrt{\pi}}{2}x)$.
$\textbf{a)}$ The phase diagram dependent on $\sigma_{w}^{2}$ and
$g$. The equilibrium feature distance $\mu(\boldsymbol{X})$ obtained
from computing the GCN GP prior for $L=4,000$ layers is shown as
a heatmap, the red line is the theoretical prediction for the transition
to the non-oversmoothing phase. $\textbf{b)}$ Same as in a) but color
coding shows whether $\mu(\boldsymbol{X})$ is close to zero (black)
or not (white) with precision $10^{-5}$. The red line again shows
the theoretically predicted phase transition. $\textbf{c)}$ Feature
distance $\mu(\boldsymbol{X}^{(l)})$ for a random input $X_{\alpha i}^{(0)}\protect\overset{\mathrm{i.i.d.}}{\sim}\mathcal{N}(0,1)$
as a function of layer $l$. Parameters are written in the panel in
matching colors and marked with color coded crosses in the phase diagram
in panel b). Feature dimension of the hidden layers is $d_{l}=200$,
crosses show the mean of $50$ network realizations, solid curves
the theoretical predictions.}
\end{figure}
To illustrate the implications of the analysis described above, we
first consider a particularly simple GCN on a complete graph; this
allows us to calculate the condition for the transition to chaos analytically,
and gain some insight into the interesting parameter regimes. Moreover,
we use this pedagogical example to show that although the GP equivalence
is only true in the limit of infinite hidden feature dimensions, $d_{l}\to\infty$,
our results still describe finite-size GCNs well.

For a complete graph with adjacency matrix $\mathcal{A}_{\alpha\beta}=1-\delta_{\alpha\beta}$,
our choice of shift operator $A$ in (\ref{eq:shift-operator-definition})
has entries $A_{\alpha\beta}=\frac{g}{N-1}+\delta_{\alpha\beta}(1-\frac{Ng}{N-1})$.
This model is a worst-case scenario for oversmoothing, since the adjacency
matrix leads to inputs that are shared across all nodes of the network.
We make the ansatz that the equilibrium covariance is of the form
$K_{\alpha\beta}^{\mathrm{eq}}=K_{c}^{\mathrm{eq}}+\delta_{\alpha\beta}(K_{a}^{\mathrm{eq}}-K_{c}^{\mathrm{eq}})$
due to symmetry which reduces the problem to only two variables. In
this formulation we can use similar methods as in the DNN case \cite{schoenholz_deep_2017}
to determine the non-oversmoothing condition on the l.h.s in (\ref{eq:chaos-condition})
(Details are given in Appendix~\ref{app:Investigation-of-toy-model}).

Figure \ref{fig:toy-model-phase-diagram} shows how a GCN on a complete
graph can be engineered to be non-oversmoothing by simple tuning of
the weight variance $\sigma_{w}^{2}$. Panel a) shows that increasing
the weight variance $\sigma_{w}^{2}$ or decreasing the size of the
off-diagonal elements $g$ both shift the network towards the non-oversmoothing
phase. Both parameters also increase the equilibrium feature distance
beyond the transition. The theoretical prediction for the transition
is calculated in Appendix~\ref{app:Investigation-of-toy-model} and
shown as the red line. Panel b) confirms the accuracy of this calculation.
Larger values of $g$ increase smoothing, and thus larger values of
$\sigma_{w}^{2}$ are needed to compensate. Moreover, our formalism
allows us to predict the evolution of feature distances over layers
correctly, as can be confirmed in panel c). We find again that GCNs
with parameters past the transition do not oversmooth.

\subsubsection{\label{subsec:General-graphs}General graphs}

For general graphs, the transition to the non-oversmoothing phase
given by Equation~(\ref{eq:chaos-condition}) can be determined numerically.
As a proof of concept, we demonstrate this approach for the Contextual
Stochastic Block Model (CSBM) \cite{deshpande_contextual_2018}, a
common synthetic model which allows generating a graph with two communities
and community-wise correlated features on the nodes. Pairs of nodes
within the same community have higher probability of being connected
and have feature vectors which are more strongly correlated, compared
with pairs of nodes from different communities.

\begin{figure}
\centering{}\includegraphics[width=1\linewidth]{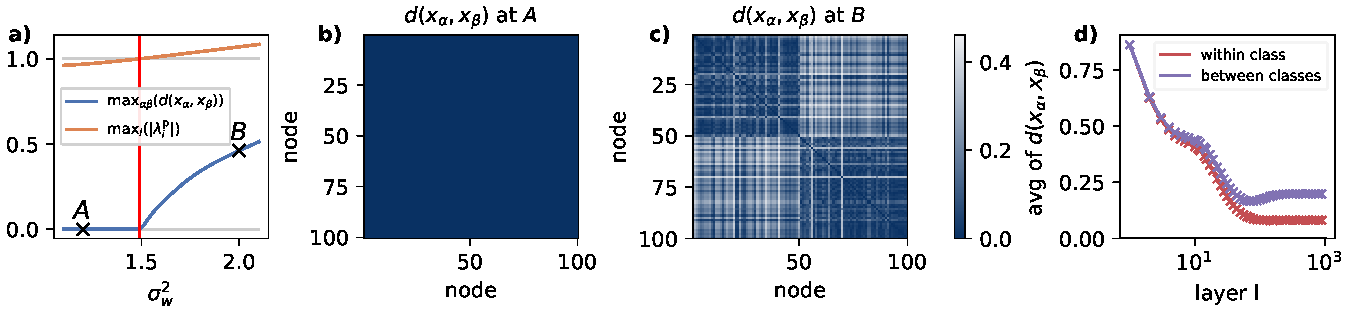}\caption{\label{fig:CSBM-non-oversmoothing-phase}The non-oversmoothing phase
in a contextual stochastic block model instance with parameters $N=100$,
$d=5$, $\lambda=1$. The shift operator is chosen according to (\ref{eq:shift-operator-definition})
with $g=0.3$, and $\sigma_{b}^{2}=0$ and $\phi(x)=\mathrm{erf}(\frac{\sqrt{\pi}}{2}x)$.
$\textbf{a)}$ The maximum feature distance between any pair of nodes
in equilibrium obtained from computing the GCN GP prior for $L=4,000$
layers (blue) and the largest eigenvalue of the linearized GCN GP
dynamics at the zero distance state as a function of weight variance
$\sigma_{w}^{2}$. The red line marks the point where $\max_{i}\{|\lambda_{i}^{\mathrm{p}}|\}=1$.
$\textbf{b)}$ Heatmap of the equilibrium distance matrix with entries
$d_{\alpha\beta}=d(\boldsymbol{x}_{\alpha},\boldsymbol{x}_{\beta})$
(Equation~(\ref{eq:2-distance-definition})) at $\sigma_{w}^{2}=1.3$,
marked as point $A$ in panel a). Colorbar shared with the plot in
c). $\textbf{c)}$ Same as b) but at point $B$ with $\sigma_{w}^{2}=2$.
$\textbf{d)}$ Features distances $d_{\alpha\beta}^{(l)}=d(\boldsymbol{x}_{\alpha}^{(l)},\boldsymbol{x}_{\beta}^{(l)})$
as a function of layers for random inputs $X_{\alpha i}^{(0)}\protect\overset{\mathrm{i.i.d.}}{\sim}\mathcal{N}(0,1)$
and a finite-size GCN with $d_{l}=200$, averaged for distances for
pairs of nodes within the same community (red) and across communities
(purple).}
\end{figure}
Given the underlying graph structure, we can construct the linear
map $H$ from Equation~(\ref{eq:linearized-gcngp}) and the analytical
solution for $C_{\theta\phi}$ in Appendix~\ref{app:Analytical-solution-for}.
Finding the set of eigenvalues is then a standard task. We show the
applicability of our formalism in Figure~\ref{fig:CSBM-non-oversmoothing-phase}
by showing that GCNs degenerate to a zero distance state state exactly
when $\sigma_{w}^{2}<\sigma_{w,\mathrm{crit}}^{2}$. Panel a) shows
how this procedure correctly predicts the transition in the given
CSBM instance: The maximum feature distance between any pair of nodes
increases from zero at the point where the state $K_{\alpha\beta}^{(l)}=k^{(l)}$
becomes unstable. This means that beyond this point, the GCN has feature
vectors that differ across nodes and therefore does not oversmooth.
This is more explicitly shown in panels b) and c), where the equilibrium
feature distance is plotted as a heatmap. At point $A$ (panel b)),
within the oversmoothing phase, all equilibrium feature distances
are indeed zero, the network therefore converges to a state in which
all features are the same. At point $B$ (panel c)) on the other hand,
pairs of nodes exist that have finite distance. In the latter case,
one can recognize the community structure of the CSBM: the lower left
and upper right quadrants are lighter than the diagonal ones, indicating
larger feature distances across communities than within. The equilibrium
state thus contains information about the graph topology. This phenomenon
is also observed in panel d), where we show the predicted feature
distance averaged for nodes within or between classes as a function
of layers compared to finite-size simulations. Again, theoretical
predictions match with simulations. Thus also on more general graphs
the presented formalism predicts the transition point between the
oversmoothing and the non-oversmoothing phase, corresponding to a
transition between regular and chaotic behavior.

We discuss how the assumptions on weight matrices in related theoretical
work \cite{cai_note_2020,wu_demystifying_2023} exclude networks in
the chaotic phase in Appendix~\ref{app:Restriction-in-related-work},
explaining why the non-oversmoothing phase has not been reported before.
In Appendix~\ref{app:KW-shift-operator} we observe how increasing
the weight variance increases the oversmoothing measure $\mu(X)$
in equilibrium also in the case of the more common shift operator
proposed in the original work \cite{kipf_semi-supervised_2017}, despite
the fact that this shift operator does not have the oversmoothed fixed
point in the sense of Equation~(\ref{eq:GCNGP-oversmoothed-fixed-point}).

\subsection{Implications for performance}

\begin{figure}
\centering{}\includegraphics[width=1\linewidth]{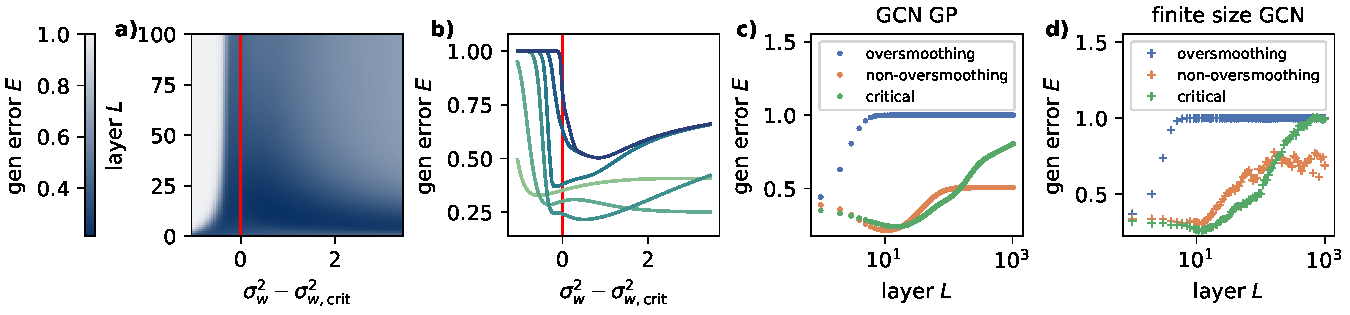}\caption{\label{fig:Generalization-error-of}Generalization error (mean squared
error) of the Gaussian process for a CSBM with parameters $N=20$,
$d=5$, $\lambda=1$, $\gamma=1$ and $\mu=4$. The shift operator
is defined in (\ref{eq:shift-operator-definition}) with $g=0.1$,
other parameters are $\sigma_{b}^{2}=0$, $\phi(x)=\mathrm{erf}(\frac{\sqrt{\pi}}{2}x)$
and $\sigma_{ro}=0.01$. In all panels we use $N^{\mathrm{train}}=10$
training nodes and $N^{\mathrm{test}}=10$ test nodes, five training
nodes from each of the two communities. Labels are $\pm1$ for the
two communities, respectively. For all panels, we show averages over
$50$ CSBM instances. $\textbf{a)}$ Heatmap of the generalization
error of the GCN GP dependent on number of layers $L$ and weight
variance $\sigma_{w}^{2}$. The red line shows the transition to the
non-oversmoothing phase. $\textbf{b)}$ Generalization error dependent
on weight variance $\sigma_{w}^{2}$ and depths $L=1,4,16,64,256,1024$
from turquoise to dark blue. $\textbf{c)}$ Generalization error dependent
on the layer for the GCN GP at the critical line $\sigma_{w}^{2}=\sigma_{w,\mathrm{crit}}^{2}$,
in the oversmoothing phase $\sigma_{w}^{2}=\sigma_{w,\mathrm{crit}}^{2}-1$
and the non-oversmoothing phase $\sigma_{w}^{2}=\sigma_{w,\mathrm{crit}}^{2}+1$.
$\textbf{d)}$ Performance of randomly initialized finite-size GCNs
with $d_{l}=200$ for $l=1,\dots,L$ where only the linear readout
layer is trained with gradient descent (details in Appendix~\ref{app:Numerical-experiments})
at the critical line $\sigma_{w}^{2}=\sigma_{w,\mathrm{crit}}^{2}$,
in the oversmoothing phase $\sigma_{w}^{2}=\sigma_{w,\mathrm{crit}}^{2}-1$
and the non-oversmoothing phase $\sigma_{w}^{2}=\sigma_{w,\mathrm{crit}}^{2}+1$.}
\end{figure}

Lastly we want to investigate the implications of the non-oversmoothing
phase on performance. We do this by applying the GCN GP as well as
a finite-size GCN to the task of node classification in the CSBM model
and measure their performance, shown in Figure~\ref{fig:Generalization-error-of}.
Panel a) shows how the generalization error of the GCN GP changes
depending on the weight variance $\sigma_{w}^{2}$ and the number
of layers $L$. In the oversmoothing phase where most GCNs in the
literature are initialized (see Appendix~\ref{app:Restriction-in-related-work}),
the generalization error increases significantly already after only
a couple of layers. We observe the best performance near the transition
to chaos where the GCN GP stays informative up to $100$ layers. In
panel b) we test the generalization error for even deeper networks.
While the generalization error increases to one (being random chance)
in the oversmoothing phase, GCN GPs in the chaotic phase stay informative
even at more than a thousand layers. This can be explained by Figure~\ref{fig:CSBM-non-oversmoothing-phase}:
For such deep networks, the dynamics are very close to the equilibrium
and thus no information of the input features $\boldsymbol{X}^{(0)}$
is transferred to the output. The state, however, still contains information
of the network topology from the adjacency matrix, leading to better
than random chance performance. In panel c) we explicitly show the
layer dependence of the generalization error for the GCN GP at the
critical point, in the oversmoothing and in the chaotic phase. Again,
we see a fast performance drop for oversmoothing networks, while in
the chaotic phase and at the critical point the GCN GP obtains good
performance also at large depths, with performance peaking at $L\approx15$
layers. Tuning the weight variance thus not only prevents oversmoothing,
but may also allow the construction of GCNs with more layers and possibly
better generalization performance.

In the study of deep networks, results obtained in the limit of infinite
feature dimensions $d_{l}\to\infty$ often are also applicable for
finite-size networks \cite{poole_exponential_2016,schoenholz_deep_2017}.
In panel d) we conduct a preliminary analysis for finite-size GCNs
by measuring the performance of randomly initialized GCNs for which
we only train the readout layer via gradient descent. Indeed, we observe
similar behavior as for the GCN GP: Performance drops rapidly over
layers in the oversmoothing phase, while performance stays high over
many layers at the critical point and in the chaotic phase and peaks
at $L\approx15$ layers.

\begin{wrapfigure}{o}{0.5\columnwidth}%
\begin{centering}
\includegraphics[width=0.5\textwidth]{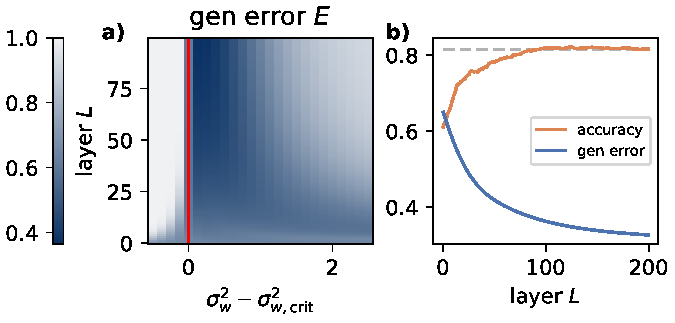}
\par\end{centering}
\caption{\label{fig:GCN-GP-performance-cora}GCN GP performance on the Cora
datset \cite{sen_collective_2008}. $\textbf{a)}$ Generalization
error (mean squared error) as a function of layers $L$ and weight
variance $\sigma_{w}^{2}-\sigma_{w,\text{crit.}}^{2}$ for our stochastic
shift operator (\ref{eq:shift-operator-definition}) with $g=0.9$.
The value of $\sigma_{w,\text{crit.}}^{2}\approx1$ is determined
numerically in Appendix~\ref{app:Non-oversmoothing-transition-cora}.
$\textbf{b)}$ Layer dependent generalization error and accuracy for
GCNs near the transition $\sigma_{w}^{2}=\sigma_{w,\text{crit.}}^{2}+0.1$.
Grey dashed line shows accuracy obtained for GCNs in the original
work \cite{kipf_semi-supervised_2017}. Numerical details in Appendix~\ref{app:Numerical-experiments}.}
\end{wrapfigure}%
Additionally, we test the performance of the GCN GP on the real world
citation network Cora \cite{sen_collective_2008}. Evaluating the
eigenvalue condition (\ref{eq:chaos-condition}) would be computationally
expensive for such a large dataset, therefore we find the transition
by numerically evaluating the feature distance $\mu(X)$ in equilibrium
and search for the $\sigma_{w}^{2}$ at which this distance becomes
non-zero. This procedure results in $\sigma_{w,\text{crit}}^{2}\approx1$
and is presented in more detail in Appendix~\ref{app:Non-oversmoothing-transition-cora}.
Figure~\ref{fig:GCN-GP-performance-cora} panel a) shows the performance
of the GCN GP dependent on the number of layers $L$ and weight variance
$\sigma_{w}^{2}$: as for the CSBM in Figure~\ref{fig:Generalization-error-of}
we observe that the performance for deep GCN GPs is best near the
transition in the non-oversmoothing phase. Furthermore, GCN GPs with
more layers achieve lower generalization error. This is shown more
directly in panel b). There we observe the layer dependence for GCN
GPs near the transition in the non-oversmoothing regime. Indeed, the
accuracy increases up to a hundred layers, reaching the accuracy of
finite-size GCNs stated in \cite{kipf_semi-supervised_2017}.

Near the transition, accuracy increases for up to $L=100$, and the
generalization error improves even beyond this. We hypothesize that
this many layers are required for high performance partly due to our
choice of the shift operator. The Cora dataset has a maximum degree
$d_{\text{max}}=168$ leading to small off-diagonal elements for the
choice of our shift operator: Recall that in Equation~(\ref{eq:shift-operator-definition}),
the parameter $g$ is constrained to be $g\in(0,1)$. As a consequence,
the off-diagonal elements of the shift operator are $A_{ij}<\frac{1}{d_{\text{max}}}=\frac{1}{168}$.
Many convolutional layers are then needed to incorporate information
from a node's neighbors.

One might wonder whether it is possible to initialize weights in say
the oversmoothing regime, and transition to the non-oversmoothing
regime during training. We argue that this is possible in the case
of Langevin training (Appendix~\ref{app:Transitioning-between-regimes}).

\section{Discussion}

In this study we used the equivalence of GCNs and GPs to investigate
oversmoothing, the property that features at different nodes converge
over layers to the same feature vector in an exponential manner. By
extending concepts such as the propagation depth and chaos from the
study of conventional deep feedforward neural networks \cite{schoenholz_deep_2017},
we are able to derive a condition to avoid oversmoothing. This condition
is phrased in terms of an eigenvalue problem of the linearized GCN
GP dynamics around the state where all features are the same: This
state is stable if all eigenvalues are smaller than one, thus the
networks do oversmooth. If one eigenvalue, however, is larger than
one, the state where the features are the same on all nodes becomes
unstable. While most GCNs studied in the literature are in the oversmoothing
phase \cite{cai_note_2020,wu_demystifying_2023}, the non-oversmoothing
phase can be reached by a simple tuning of the weight variance at
initialization. An analogy can be drawn between the chaotic phase
of DNNs and the non-oversmoothing phase of GCNs. Previous theoretical
works have proven that oversmoothing is inevitable in some GNN architectures,
among them GCNs; these works, however, make crucial assumptions on
the weight matrices, constraining their variances to be in what we
identify as the oversmoothing phase. Near the transition, we find
GCNs which are both deep and expressive, matching the originally reported
GCN performance \cite{kipf_semi-supervised_2017} on the Cora dataset
with GCN GPs beyond $100$ layers.

$\textbf{Limitations.}$ The current analysis is based on the equivalence
of GCNs and GPs which strictly holds only in the limit of infinite
feature dimension. GCNs with large feature vectors ($d_{l}=200$)
are well described by the theory, as shown in Section~\ref{sec:Results}.
For a small number of feature dimensions, however, we expect deviations
from the asymptotic results. Throughout the main part of this work,
we assumed a row-stochastic shift operator which made the equilibrium
$K^{\mathrm{eq}}$ in the oversmoothing phase particularly simple.
For other shift operators, we expect qualitatively similar results
while the equilibrium $K^{\mathrm{eq}}$ may look different in detail.
In our preliminary experiments on the common shift operator from \cite{kipf_semi-supervised_2017}
(Appendix~\ref{app:KW-shift-operator}), we indeed find that increasing
the weight variance increases the distances between features also
in this case. We hypothesize that this effect makes the equilibrium
more informative of the graph topology, as in the stochastic shift
operator case. The choice of non-linearity is unrestricted, but in
the general case numerical integration of (\ref{eq:GCNGP_expectation_value})
is needed.

To determine whether a given weight variance is in the non-oversmoothing
phase, one calculates the eigenvalues of the linearized GCN GP dynamics
which take the form of an $N^{2}\times N^{2}$ matrix (see Equation~(\ref{eq:linearized-gcngp})),
this has a run time of $\mathcal{O}(N^{6})$. While this becomes computationally
expensive for large graphs, the conceptual insights of the presented
analysis remain. In practical applications with large graphs one may
reduce the computational load by determining the transition point
via computation of the GCN GP prior until it is close to equilibrium.
This procedure has a runtime of $\mathcal{O}(N^{3}L^{\mathrm{eq}})$
where $L^{\mathrm{eq}}$ is the number of layers after which the process
is sufficiently close to equilibrium. One might then do an interval
search on the weight variance until the transition point is determined
with sufficient accuracy.

$\textbf{Oulook.}$ Formulating GCNs with the help of GPs can be considered
the leading order in the number of feature space dimension $d_{l}$
when approximating finite-size GCNs. Computing corrections for finite
numbers of hidden feature dimensions would allow the characterization
of feature learning in such networks, similar as in standard deep
networks \cite{naveh_self_2021,zavatone-veth_contrasting_2022,seroussi_separation_2023}.
Moreover, the generalization of this formalism to more general shift
operators and other GNN architectures \cite{zhao_pairnorm_2020,chen_simple_2020}
like GATs \cite{velickovic_graph_2018} are possible directions of
future research. In the special case of GATs we expect similar results
to the GCN analyzed here, since the shift operator is constructed
using a softmax and therefore also is row-stochastic.

\subsection*{Acknowledgements}

Funded by the European Union (ERC, HIGH-HOPeS, 101039827). Views and
opinions expressed are however those of the author(s) only and do
not necessarily reflect those of the European Union or the European
Research Council Executive Agency. Neither the European Union nor
the granting authority can be held responsible for them. We also acknowledge
funding by the German Research Council (DFG) within the Collaborative
Research Center “Sparsity and Singular Structures” (SfB 1481; Project
A07).

 \newpage{}

\appendix

\section{Analytical solution for expectation values}

\label{app:Analytical-solution-for}To evaluate our theory, we need
to compute expectation values given in the form of (\ref{eq:GCNGP_expectation_value})
which we restate here for readability
\begin{align}
C_{\gamma\delta}^{(l)} & =\Big\langle\phi\Big(h_{\gamma}^{(l)}\Big)\phi\Big(h_{\delta}^{(l)}\Big)\Big\rangle_{h_{\gamma}^{(l)},h_{\delta}^{(l)}}\,,
\end{align}
where $h_{\gamma}^{(l)}$ and $h_{\delta}^{(l)}$ are zero mean random
Gaussian variables with $\langle h_{\gamma}^{(l)}h_{\delta}^{(l)}\rangle=K_{\gamma\delta}^{(l)}$.
This can be evaluated numerically for general non-linearities $\phi(x)$.
For simplicity, however, we choose $\phi(x)=\mathrm{erf}(\frac{\sqrt{\pi}}{2}x)$
in our experiments where the scaling factor in the $\mathrm{erf}$
is chosen such that $\frac{\partial\phi}{\partial x}(0)=1$. In this
case, the expectation value can be evaluated analytically to be 
\begin{align}
C_{\gamma\delta}^{(l)} & =\frac{2}{\pi}\,\arcsin\Bigg(\frac{\frac{\pi}{2}K_{\gamma\delta}^{(l)}}{\sqrt{1+\frac{\pi}{2}K_{\gamma\gamma}^{(l)}}\sqrt{1+\frac{\pi}{2}K_{\delta\delta}^{(l)}}}\Bigg)\,.
\end{align}
as shown in \cite{williams_computation_1998}.

\section{The linearized GP of GCNs}

\label{app:Linearized-Gaussian-process}We start from the full GCN
GP iterative map (\ref{eq:GCNGP_expectation_value},(\ref{eq:GCNGP_K_update})),
which we restate here for readability
\begin{align}
K_{\alpha\beta}^{(l+1)}= & T_{\alpha\beta}[K^{(l)}]=\sigma_{b}^{2}+\sigma_{w}^{2}\,\sum_{\gamma,\delta}A_{\alpha\gamma}A_{\beta\delta}C_{\gamma\delta}^{(l)}[K^{(l)}]
\end{align}
where $h_{\gamma}^{(l)}$ and $h_{\delta}^{(l)}$ are drawn from a
$0$-mean Gaussian distribution with covariance $\langle h_{\gamma}^{(l)}h_{\delta}^{(l)}\rangle=K_{\gamma\delta}^{(l)}$.
The full covariance matrix of layer $l$ is denoted as $K^{(l)}$.
Here, we distinguish between the iterative maps $T_{\alpha\beta}:\mathbb{R}^{N\times N}\to\mathbb{R}$
of which there are $N^{2}$, one for each pair of nodes $\alpha$
and $\beta$, and the entries $K_{\alpha\beta}^{(l+1)}$ of the covariance
matrix in the layer $l+1$. Notice that $T_{\alpha\beta}=T_{\beta\alpha}$
due to the symmetry of the covariance matrix. In the maps $T_{\alpha\beta}$,
the covariance matrix of the previous layer only shows up in the expectation
value $C_{\gamma\delta}=\Big\langle\phi\Big(h_{\gamma}^{(l)}\Big)\phi\Big(h_{\delta}^{(l)}\Big)\Big\rangle_{h_{\gamma}^{(l)},h_{\delta}^{(l)}}$
such that the linearized dynamics around a fixed point (being equilibrium
$K_{\alpha\beta}^{\mathrm{fix}}=K_{\alpha\beta}^{\mathrm{eq}}$ or
zero distance state $K_{\alpha\beta}^{\mathrm{fix}}=k$) with $K_{\alpha\beta}^{(l)}=K_{\alpha\beta}^{\mathrm{fix}}+\Delta_{\alpha\beta}^{(l)}$
read
\begin{align}
K_{\alpha\beta}^{(l+1)}=K_{\alpha\beta}^{\mathrm{fix}}+\Delta_{\alpha\beta}^{(l+1)}= & \underbrace{T_{\alpha\beta}[K^{\mathrm{fix}}]}_{=K_{\alpha\beta}^{\mathrm{fix}}}+\sum_{\gamma<\delta}\frac{\partial T_{\alpha\beta}}{\partial K_{\gamma\delta}}[K^{\mathrm{fix}}]\Delta_{\gamma\delta}^{(l)}+\mathcal{O}((\Delta^{(l)})^{2})\label{eq:linearization_restricted_sum}\\
= & K_{\alpha\beta}^{\mathrm{fix}}+\sum_{\gamma,\delta}\underbrace{\sigma_{w}^{2}\sum_{\theta,\phi}\frac{1}{2}(1+\delta_{\gamma,\delta})A_{\alpha\theta}A_{\beta\phi}\frac{\partial C_{\theta\phi}}{\partial K_{\gamma\delta}}[K^{\mathrm{fix}}]}_{\equiv H_{\alpha\beta,\gamma\delta}}\Delta_{\gamma\delta}^{(l)}+\mathcal{O}((\Delta^{(l)})^{2})\,,
\end{align}
where we restrict the sum in (\ref{eq:linearization_restricted_sum})
to $\gamma<\delta$ since $K_{\gamma\delta}$ and $K_{\delta\gamma}$
are the same quantity. From (\ref{eq:linearization_restricted_sum})
follows
\begin{align}
\Delta_{\alpha\beta}^{(l+1)}= & \sum_{\gamma,\delta}H_{\alpha\beta,\gamma\delta}\Delta_{\gamma\delta}^{(l)}+\mathcal{O}((\Delta^{(l)})^{2})
\end{align}
which is Equation~(\ref{eq:linearized-gcngp}) in the main text.
While $H$ is not symmetric in general, $H_{\alpha\beta,\gamma\delta}\neq H_{\gamma\delta,\alpha\beta}$,
it is symmetric in the first and second pair of covariance indices,
$H_{\alpha\beta,\gamma\delta}=H_{\beta\alpha,\gamma\delta}$ and $H_{\alpha\beta,\gamma\delta}=H_{\alpha\beta,\delta\gamma}$
due to symmetry of the covariance matrices, $K_{\alpha\beta}=K_{\beta\alpha}$.

In the main text, we look for the right eigenvectors of $H$ fulfilling
\begin{align}
\lambda_{i}V_{\alpha\beta}^{(i)}= & \sum_{\gamma,\delta}H_{\alpha\beta,\gamma\delta}V_{\gamma\delta}^{(i)}\,.
\end{align}
These are for general non-symmetric matrices not orthogonal. In order
to decompose $\Delta^{(l)}$ to overlaps with the eigenvectors $V_{\alpha\beta}^{(i)}$
we need to find the dual vectors $U_{\alpha\beta}^{(i)}$ fulfilling
\begin{align}
\sum_{\alpha,\beta}U_{\alpha\beta}^{(i)}V_{\alpha\beta}^{(j)}= & \delta_{ij}
\end{align}
from which we can define
\begin{align}
\Delta_{i}^{(l)}= & \sum_{\alpha,\beta}U_{\alpha\beta}^{(i)}\Delta_{\alpha\beta}^{(l)}
\end{align}
such that
\begin{align}
\Delta_{\alpha\beta}^{(l)}= & \sum_{i}\Delta_{i}^{(l)}V_{\alpha\beta}^{(i)}
\end{align}
as stated in the main text.

\section{Investigation of the complete graph model}

\label{app:Investigation-of-toy-model}In this section we analytically
investigate the complete graph model as defined in the main text.
Specifically, we consider networks with the shift operator
\begin{align}
A_{\alpha\beta} & =\frac{g}{N-1}+\delta_{\alpha\beta}(1-\frac{Ng}{N-1})\label{eq:toy_model_shift_operator}
\end{align}
and $\phi=\mathrm{erf}(\sqrt{\pi}x/2)$. Due to the symmetry of the
system we make the ansatz
\begin{align}
K_{\alpha\beta}^{\mathrm{eq}} & =K_{\mathrm{a}}^{\mathrm{eq}}+\delta_{\alpha\beta}(K_{\mathrm{c}}^{\mathrm{eq}}-K_{\mathrm{a}}^{\mathrm{eq}})\,,
\end{align}
meaning that we assume constant variances $K_{\mathrm{a}}^{\mathrm{eq}}$
across nodes and that all pairs of nodes have the same covariance
$K_{\mathrm{c}}^{\mathrm{eq}}$, reducing the system to only two unknown
variables. The equilibrium is a fixed point of the iterative map of
the GCN GP. With the special choice of shift operator in (\ref{eq:toy_model_shift_operator})
this becomes
\begin{align}
K_{\mathrm{a}}^{(l+1)} & =\sigma_{b}^{2}+g_{\mathrm{a}}C_{\mathrm{a}}^{(l)}+g_{\mathrm{c}}C_{\mathrm{c}}^{(l)}
\end{align}
and
\begin{align}
K_{\mathrm{c}}^{(l)} & =\sigma_{b}^{2}+h_{\mathrm{a}}C_{\mathrm{a}}^{(l)}+h_{'\mathrm{c}}C_{\mathrm{c}}^{(l)}
\end{align}
with constants
\begin{align}
g_{\mathrm{a}} & =(1+\frac{g^{2}}{N-1})\sigma_{w}^{2}\\
g_{\mathrm{c}} & =2(g+\frac{g^{2}(N-2)}{N-1})\sigma_{w}^{2}\\
h_{\mathrm{a}} & =2(\frac{g}{N-1}+\frac{g^{2}(N-2)}{(N-1)^{2}})\sigma_{w}^{2}\\
h_{\mathrm{c}} & =(1+\frac{g^{2}}{(N-1)^{2}}+2\frac{g^{2}(N-2)(N-3)}{(N-1)^{2}}+4\frac{g(N-2)}{(N-1)})\sigma_{w}^{2}
\end{align}
and
\begin{align}
C_{\mathrm{a}}^{(l)} & =\langle\phi(h_{\alpha}^{(l)})\phi(h_{\alpha}^{(l)})\rangle\\
C_{\mathrm{c}}^{(l)} & =\langle\phi(h_{\alpha}^{(l)})\phi(h_{\beta}^{(l)})\rangle\qquad\text{for }\alpha\neq\beta\,.
\end{align}
The preactivations are Gaussian distributed with zero mean and covariance
$\langle h_{\alpha}^{(l)}h_{\beta}^{(l)}\rangle=K_{\alpha\beta}^{(l)}$.
In the oversmoothing phase, we know that $\mu(\boldsymbol{X})=0$
in equilibrium. We have seen in Section~\ref{subsec:The-non-oversmoothing-phase}
that this corresponds to $K_{\alpha\beta}^{\mathrm{eq}}=k^{\mathrm{eq}}$,
implying for our ansatz that $K_{\mathrm{c}}^{\mathrm{eq}}=K_{\mathrm{a}}^{\mathrm{eq}}$.
We will find the transition to chaos by calculating where this state
becomes unstable with regard to small perturbations. Specifically,
we define $c^{(l)}=\frac{K_{\mathrm{c}}^{(l)}}{K_{\mathrm{a}}^{(l)}}$
and look for the parameter point where
\begin{align}
1 & \overset{?}{>}\frac{\partial c^{(l+1)}}{\partial c^{(l)}}\Big|_{c^{(l)}=1}\,.\label{eq:appendix-chaos-condition-toy-model}
\end{align}
The authors in \cite{schoenholz_deep_2017} used this approach to
find the transition to chaos for DNNs. The correlation coefficient
is
\begin{align}
c^{(l+1)} & =\frac{K_{\mathrm{c}}^{(l+1)}}{K_{\mathrm{a}}^{(l+1)}}=\frac{\sigma_{b}^{2}+h_{\mathrm{a}}C_{\mathrm{a}}^{(l)}+h_{\mathrm{c}}C_{\mathrm{c}}^{(l)}}{\sigma_{b}^{2}+g_{\mathrm{a}}C_{\mathrm{a}}^{(l)}+g_{\mathrm{c}}C_{\mathrm{c}}^{(l)}}
\end{align}
and Equation~(\ref{eq:appendix-chaos-condition-toy-model}) thus
becomes
\begin{align}
\frac{\partial c^{(l+1)}}{\partial c^{(l)}}\Big|_{c^{(l)}=1} & =\frac{h_{\mathrm{c}}\frac{\partial C_{\mathrm{c}}^{(l+1)}}{\partial c^{(l)}}K_{\mathrm{a}}^{(l+1)}-g_{\mathrm{c}}\frac{\partial C_{\mathrm{c}}^{(l+1)}}{\partial c^{(l)}}K_{\mathrm{c}}^{(l+1)}}{(K_{\mathrm{a}}^{(l+1)})^{2}}\,.\label{eq:appendix-toy-chaos-calculated}
\end{align}
Since we look at this equation at the perfectly correlated state $c^{(l)}=1$,
we know that $K_{\mathrm{a}}^{(l)}=K_{\mathrm{c}}^{(l)}$ (implying
that $C_{\mathrm{a}}^{(l)}=K_{\mathrm{c}}^{(l)}$) and can determine
$K_{\mathrm{a}}^{(l)}$ as the solution of the fixed point equation
\begin{align}
K_{\mathrm{a}}^{(l+1)} & =\sigma_{b}^{2}+(g_{\mathrm{a}}+g_{\mathrm{c}})C_{\mathrm{a}}^{(l)}
\end{align}
to which the GCN GP dynamics reduce in the zero distance state (by
using the fact that $\sum_{\beta}A_{\alpha\beta}=1$ an $C_{\alpha\beta}^{(l)}=c^{(l)}$).
Lastly, we can calculate
\begin{align}
\frac{\partial C_{\mathrm{c}}^{(l+1)}}{\partial c^{(l)}}\Big|_{c=1} & =\frac{\partial}{\partial c^{(l)}}\Bigg(\frac{2}{\pi}\arcsin\Bigg(\frac{\frac{\pi}{2}K_{\mathrm{a}}^{(l)}c^{(l)}}{1+\frac{\pi}{2}K_{\mathrm{a}}^{(l)}}\Bigg)\Bigg)\Bigg|_{c^{(l)}=1}\\
 & =\frac{2}{\pi}\frac{1}{\sqrt{1-\big(\frac{\frac{\pi}{2}K_{\mathrm{a}}^{(l)}c^{(l)}}{1+\frac{\pi}{2}K_{\mathrm{a}}^{(l)}}\big)^{2}}}\frac{K_{\mathrm{a}}^{(l)}}{\frac{2}{\pi}+K_{\mathrm{a}}^{(l)}}\Bigg|_{c^{(l)}=1}\\
 & =\frac{2}{\pi}\frac{1}{\sqrt{1-\big(\frac{K_{\mathrm{a}}^{(l)}}{\frac{2}{\pi}+K_{\mathrm{a}}^{(l)}}\big)^{2}}}\frac{K_{\mathrm{a}}^{(l)}}{\frac{2}{\pi}+K_{\mathrm{a}}^{(l)}}\,,
\end{align}
where we used the known solution for the expectation value $C_{\mathrm{c}}$
for $\phi(x)=\mathrm{erf}(\sqrt{\pi}x/2)$ from Appendix~\ref{app:Analytical-solution-for}.
Plugging this into Equation~(\ref{eq:appendix-toy-chaos-calculated})
lets us calculate $\frac{\partial c^{(l+1)}}{\partial c^{(l)}}\Big|_{c^{(l)}=1}$
and thus determine the transition to the non-oversmoothing phase.
This is plotted as a red line in Figure~\ref{fig:toy-model-phase-diagram}
panel a) and b).

\section{\label{app:Restriction-in-related-work}Restriction of weight matrices
in related work}

In this section we will show histograms of critical weight variances
$\sigma_{w,\mathrm{crit}}^{2}$ and discuss how the assumptions in
\cite{cai_note_2020} and \cite{wu_demystifying_2023} exclude networks
in the non-oversmoothing phase. Our results thus stand in no conflict
with the results of these works.

\begin{figure}
\centering{}\includegraphics[width=0.33\linewidth]{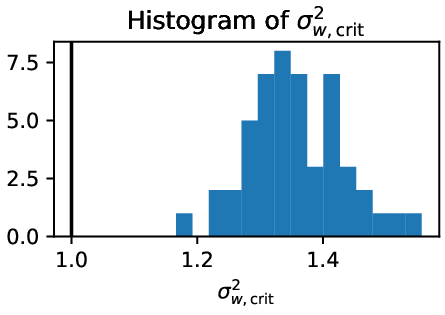}\caption{\label{fig:app-histogram-of-critvals}Histogram of $\sigma_{w,\mathrm{crit}}^{2}$
for the $50$ CSBM instances used in the experiment of Figure~\ref{fig:Generalization-error-of}.
The point $1$ is marked for comparison to related work.}
\end{figure}

Here we want to argue that the assumptions on the weight matrices
in related work constrains their architectures to the oversmoothing
phase. We start with \cite{wu_demystifying_2023} in which the authors
study graph attention networks (GATs). Although we study a standard
GCN here, we hypothesize that increasing the weight variance at initialization
likewise prevents oversmoothing in other architecture, such as the
GAT in \cite{wu_demystifying_2023}. The critical assumption constraining
them to the oversmoothing phase is their assumption A3, stating that
$\{||\prod_{l=0}^{k}|W^{(l)}|||_{\max}\}_{k=0}^{\infty}$ is bounded
where $||M||_{\max}=\max_{i,j}|M_{i,j}|$. For our setting of randomly
drawn weight matrices $W_{ij}^{(l)}\overset{\mathrm{i.i.d.}}{\sim}\mathcal{N}(0,\frac{\sigma_{w}^{2}}{d_{l}})$
with $W_{ij}^{(l)}\in\mathbb{R}^{N\times N}$, this restricts us to
values $\sigma_{w}^{2}\leq1$. This can be seen by using the circular
law from random matrix theory \cite{ginibre_statistical_1965}: It
is known that the eigenvalues of a matrix with i.i.d. random Gaussian
entries of the form above have eigenvalues uniformly distributed in
a circle around $0$ in the complex plane with radius $\sqrt{\sigma_{w}^{2}}$
in the limit $d_{l}\to\infty$. Thus, the maximal real part of any
eigenvalue of this matrix is $\sqrt{\sigma_{w}^{2}}$. Thus we can
estimate $||\prod_{l=0}^{k}|W^{(l)}|||_{\max}\leq c(\sqrt{\sigma_{w}^{2}})^{k}$
with a constant $c$. To enter the non-oversmoothing phase, we need
$\sigma_{w}^{2}>1$. In this case, the latter expression diverges
for $k\to\infty$, thus being excluded by the proof in \cite{wu_demystifying_2023}.
Indeed, for the CSBMs we investigated in this work, all critical weight
variances $\sigma_{w,\mathrm{crit}}^{2}$ are larger than $1$ as
shown in Figure~\ref{fig:app-histogram-of-critvals}. Also in our
model of the complete graph and the CSBM investigated in Section~\ref{subsec:General-graphs}
all $\sigma_{w,\mathrm{crit}}^{2}$ are larger than $1$, compare
Figure~\ref{fig:toy-model-phase-diagram} and Figure~\ref{fig:CSBM-non-oversmoothing-phase}.

The authors of \cite{cai_note_2020} and \cite{oono_graph_2021} also
study GCNs,; however they consider a different shift operator than
in this work. In both of \cite{cai_note_2020} and \cite{oono_graph_2021}
the authors find that their GCN models exponentially loose expressive
power if $s\overline{\lambda}<1$, with $\overline{\lambda}$ being
the maximal singular value/eigenvalue of the shift operator and $s$
being the maximal singular value of all weight matrices. Again, the
maximal singular value is limited (dependent on $\overline{\lambda}$),
which in our approach translates to a limit on $\sigma_{w}^{2}$.
While the authors notice that their bounds do not hold for large singular
values $s$, they do not observe a non-oversmoothing phase in their
models.

\section{\label{app:KW-shift-operator}Non-oversmoothing GCNs with non-stochastic
shift operators}

Here we investigate how increasing the weight variance can mitigate
oversmoothing also in the case of the original shift operator proposed
by Kipf and Welling in \cite{kipf_semi-supervised_2017} being
\begin{align}
A_{\text{KW}} & =(D^{\prime})^{-1/2}\mathcal{A^{\prime}}(D^{\prime})^{-1/2}\label{eq:Kipf_Welling_shift_operator}
\end{align}
with $\mathcal{A}^{\prime}=\mathbb{I}+\mathcal{A}$ and $D_{ij}^{\prime}=\delta_{ij}\sum_{k}\mathcal{A}_{ik}^{\prime}$.
This shift operator $A_{\text{KW}}$ is not row-stochastic, therefore
the state $K_{\alpha\beta}=k$ is not a fixed point for $k\neq0$,
as can be seen from Equation~(\ref{eq:GCNGP-oversmoothed-fixed-point}).
Thus we need to differentiate between two kind of fixed points: Either,
all features are zero, for which $K_{\alpha\beta}=0$ and $\mu(X)=0$,
or some $K_{\alpha\beta}\neq0$ in which case we have $\mu(X)>0$.
Importantly, for this shift operator, there is no intermediate case
for which $K_{\alpha\beta}=k$ with $k\neq0$. Consequently there
is no oversmoothing regime with respect to the measure $\mu(X)$ where
node features are non-zero. For an in depth study of this case (\ref{eq:Kipf_Welling_shift_operator}),
the definition of another oversmoothing measure $\mu^{\prime}$ incorporating
the different values of $\sum_{\beta}(A_{\text{KW}})_{\alpha\beta}$
for different $\alpha$ may be more appropriate. For our purposes,
it will suffice to analyze the shift operator (\ref{eq:Kipf_Welling_shift_operator})
with the measure $\mu(X)$ from Equation~(\ref{eq:node-sim-definition}).

\begin{figure}
\centering{}\includegraphics[width=1\linewidth]{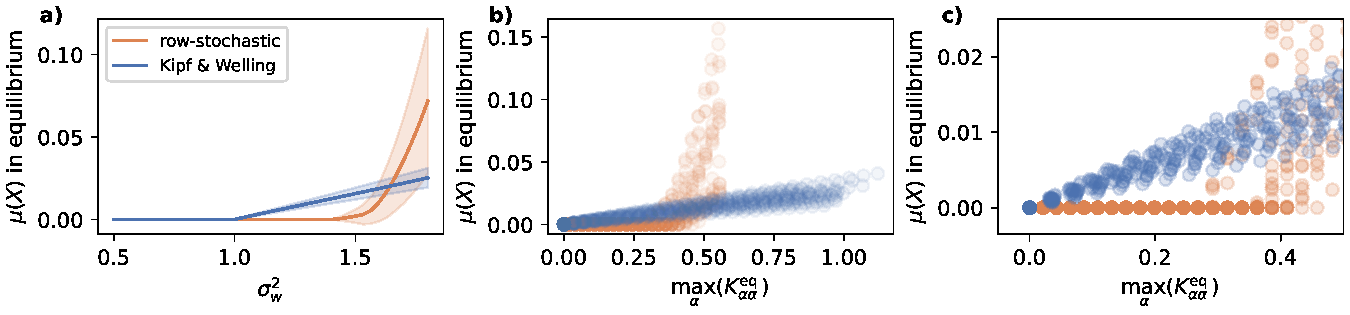}\caption{\label{fig:app-oversmoothing-Kipf-Welling-A}Oversmoothing in GCN
GPs with the commonly used shift operator (\ref{eq:Kipf_Welling_shift_operator})
(called Kipf \& Welling in the label) and our row-stochastic shift
operator (\ref{eq:shift-operator-definition}). $\textbf{a)}$ Feature
distance $\mu(X)$ in equilibrium dependent on weight variance $\sigma_{w}^{2}$
obtained from simulating the GCN GP priors for $L^{\text{eq}}=4,000$
layers. Shown are the averages (solid lines) and standard deviations
(shaded areas) over $20$ CSBM initializations with $\lambda=1$,
$d=5$ and $N=30$ nodes.$\textbf{b)}$ Scatter plot of maximal variance
of all nodes $\max_{\alpha}\{K_{\alpha\alpha}^{\text{eq}}\}$ and
feature distance $\mu(X)$ in equilibrium for the same data as in
plot a). $\textbf{c)}$ Same as b) but zoomed into lower left corner.}
\end{figure}

Figure~\ref{fig:app-oversmoothing-Kipf-Welling-A} panel a) shows
how $\mu(X)$ in equilibrium increases for larger weight variances
$\sigma_{w}^{2}$ for the shift operator (\ref{eq:Kipf_Welling_shift_operator}).
For comparison, we also show the results obtained with our row-stochastic
shift operator (\ref{eq:shift-operator-definition}). Thus we find
that also for the shift operator $A_{\text{KW}}$ the pairwise distance
between features can be increased by increasing the weight variance
$\sigma_{w}^{2}$. The non-existence of an oversmoothed fixed point
except for the special case $K_{\alpha\beta}=0$ which we argued for
above is observed in panel b) and c): The oversmoothing measure $\mu(X)$
is zero for $A_{\text{KW}}$ if and only if all entries of the covariance
matrix are zero, implying that all preactivations and thus also all
features are zero. This is a qualitative difference to our shift operator
$A$ for which we find equilibrium states with non-zero $\max_{\alpha}K_{\alpha\alpha}^{\text{eq}}$
but still $\mu(X)=0$.

\section{\label{app:Numerical-experiments}Details of numerical experiments}

To conduct our experiments we use NumPy \cite{harris_array_2020},
SciPy \cite{virtanen_scipy_2020} (both available under a BSD-3-Clause
License) and Scikit-learn (sklearn) \cite{pedregosa_scikit-learn_2011}
(available under a New BSD License). The code is publicly available
under \url{https://github.com/bepping/non-oversmoothing-gcns}. For
our experiments with the Cora dataset, we use the readin methods from
\cite{kipf_semi-supervised_2017} which are available under a MIT
license (Copyright (c) 2016 Thomas Kipf). Computations were performed
on CPUs.Requirements for the experiments with synthetic data are (approximately):
\begin{itemize}
\item Figure~\ref{fig:toy-model-phase-diagram}: 10mins on a single core
laptop.
\item Figure~\ref{fig:CSBM-non-oversmoothing-phase}: 10h on a single node
on an internal CPU cluster. Most of the computation time is needed
for evaluating $\max(\lambda_{i}^{\mathrm{p}})$ in panel a).
\item Figure~\ref{fig:Generalization-error-of}: 2h on a single core laptop.
In panel d), the last layer of finite-size GCNs is trained with the
standard settings from sklearn.linear\_model.SGDRegressor().
\item Figure~\ref{fig:app-histogram-of-critvals}: Byproduct of Figure~\ref{fig:CSBM-non-oversmoothing-phase}.
\item Figure~\ref{fig:app-oversmoothing-Kipf-Welling-A}: 10min on a single
core laptop.
\end{itemize}
We also experimented on the real world benchmark dataset Cora \cite{sen_collective_2008}.
This is a citation network with $2708$ nodes, representing publications,
and $5429$ edges, representing the citations between them (we use
undirected edges). The publications are divided into seven classes,
and the task is to predict these classes for the unlabeled nodes.
Node features are of $0$/$1$ valued vectors indicating the absence/presence
of words from a dictionary in the titles of the respective publication.
The dictionary consists of $1433$ unique words. Requirements for
the experiments with the Cora dataset are:
\begin{itemize}
\item Figure~\ref{fig:GCN-GP-performance-cora}: 1h on a single core laptop.
\item Figure~\ref{fig:app-cora-transition}: 15h on a single core laptop.
\end{itemize}

\section{\label{app:Non-oversmoothing-transition-cora}Non-oversmoothing transition
in the Cora dataset}

\begin{figure}
\centering{}\includegraphics[width=0.33\linewidth]{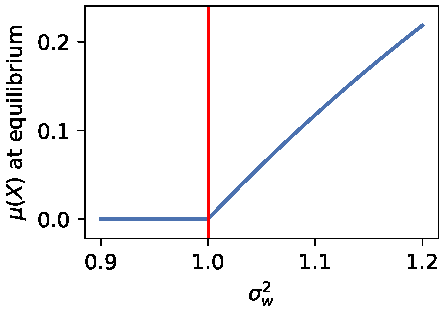}\caption{\label{fig:app-cora-transition}Node distance measure $\mu(\boldsymbol{X})$
at equilibrium obtained from computing the GCN GP prior for $L=4,000$
layers as a function of $\sigma_{w}^{2}$. The transition to the non-oversmoothing
regime is estimated by checking where the node distance measure is
larger than $\epsilon=10^{-5}$, marked as the red line.}
\end{figure}

For Figure~\ref{fig:GCN-GP-performance-cora} we numerically determined
the transition to the non-oversmoothing regime for the Cora dataset.
The transition was estimated by measuring the distance $\mu(\boldsymbol{X})$
at equilibrium (i.e., after many layers), and determining at which
value of $\sigma_{w}^{2}$ it becomes larger than a small distance
$\epsilon=10^{-5}$. The results of this experiment are shown in Figure~\ref{fig:app-cora-transition}.
We find the critical point to be $\sigma_{w,\text{crit}}^{2}=1$ while
using a step size of $\delta\sigma_{w}^{2}=0.01$.

\section{\label{app:Transitioning-between-regimes}Transitioning between regimes
during training}

In this section argue why we think it is possible to transition from
the oversmoothing to the non-oversmoothing regime or vice versa during
training.

In the GP limit of infinitely many hidden features with a finite amount
of training data, the variance of weights of a neural network is the
same before and after training. This is because weights only change
marginally in this limit, also known as the lazy training regime \cite{jacot_neural_2018,geiger_disentangling_2020}.
We will use this fact to argue that Langevin training is capable of
transitioning from one regime to the other.

Langevin training is a gradient based training scheme with external
noise and a decay term. The gradient flow equation is given as
\[
\frac{\mathrm{d}W_{ij}}{\mathrm{d}t}=-\gamma W_{ij}-\nabla_{W_{ij}}L+B_{ij}\,,
\]
where $W_{ij}$ are the weights to be trained, $L$ is the loss function
and $B_{ij}$ denotes external white noise $\langle B_{ij}(t)B_{kl}(s)\rangle=\delta_{i,k}\delta_{j,l}\delta(t-s)D$
with strength $D$ where $\delta_{a,b}$ and $\delta(a-b)$ denote
the Kronecker or Dirac delta, respectively. Both $\gamma$ and $D$
are parameters of this training scheme. It is known that the distribution
of weights converges to the posterior weight distribution of a neural
network GP with weight variance $\sigma_{w}^{2}=\frac{D}{2\gamma}$
in the infinite feature limit \cite{naveh_predicting_2021} which,
as we have noticed above, is the same as the prior distribution.

Since the Langevin distribution converges to the GP weight posterior
for any initial distribution, one can imagine an initial distribution
with a variance that initializes the network in the oversmoothing
regime, while the parameters $\gamma$ and $D$ are chosen such that
$\sigma_{w}^{2}=\frac{D}{2\gamma}>\sigma_{w,\text{crit}}^{2}$ implying
that after training most weight realizations will be in the non-oversmoothing
regime. Thus, in this case the GCN would have transitioned from one
regime to the other. While this argument is made in the limit of infinitely
many hidden features, we think that qualitatively similar results
are possible for a large but finite number of hidden feature dimensions.
The transition in the reverse direction is possible by the same argument
only with the initial and the final variances exchanged.
\end{document}